# Text Line Identification in Tagore's Manuscript

Chandranath Adak
Dept. of Computer Science & Engg.
University of Kalyani
West Bengal-741235, India
adak32@gmail.com

Bidyut B. Chaudhuri
Computer Vision and Pattern Recognition Unit
Indian Statistical Institute
Kolkata-700108, India
bbc@isical.ac.in

*Abstract*—In this paper, a text line identification method is proposed. The text lines of printed document are easy to segment due to uniform straightness of the lines and sufficient gap between the lines. But in handwritten documents, the line is non-uniform and interline gaps are variable. We take Rabindranath Tagore's manuscript as it is one of the most difficult manuscripts that contain doodles. Our method consists of a preprocessing stage to clean the document image. Then we separate doodles from the manuscript to get the textual region. After that we identify the text lines on the manuscript. For text line identification, we use window examination, black run-length smearing, horizontal histogram and connected component analysis.

## I. INTRODUCTION

In the field of document image analysis [1], the historical documents are extremely important. Some historical documents are priceless, rare and collected by a great effort. Such documents are usually preserved in the museum. In our work, we are concerned on manuscript image processing where we deal with handwritten [2] documents (manuscripts) of Nobel laureate *Rabindranath Tagore (1861-1941)*.

We choose Tagore's manuscript because it is one of the difficult manuscripts. If we are successful with Tagore's manuscript then it may be possible to handle more tough cases. Tagore's manuscript makes it difficult due to presence of non-text doodles (*fig.1*) of real and imaginary characters, animals, human, trees and phantoms. In the way of making an OHCR (Optical Handwritten Character Recognizer) [4] on Tagore's handwritten manuscript, it is challenging to identify the textual region [3] on the manuscript.

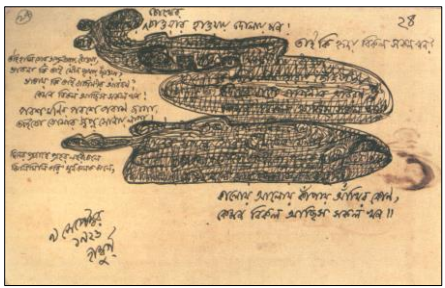

Fig.1. Manuscript of Tagore's song ("Chokher chaoar haoay dolay mon"), *Geetabitan*, 9 Sep., 1926, contains a doodle.

In this paper, we propose to identify the text lines on Tagore's manuscript. Our method contains a pre-processing stage, where *median filter* is used for smoothing and *Fuzzy C-Means* (FCM) clustering is used for segmentation into foreground (ink) and background (paper). Here in §.II on the proposed method, we explain the further steps of text/non-text separation and text line identification. We obtained 89.41% of F-measure on the basis of our experiment on the test cases, presented in §.III.

## II. PROPOSED METHOD

The proposed method is as follows:

### A. Preprocessing

The Tagore's manuscripts that we considered are very old. So the possibility of quality degradation is high, and there exists distortion due to ink-stain and bleed through. The threshold based binarization does not work well to deal in this case. We use *FCM* clustering to find our region of interest (ink-strokes) and find satisfactory outcomes.

At first a *3X3* median filter is used to smooth the manuscript image. Next *FCM* clustering is applied on this image to segment it into foreground (ink-strokes) and background (paper). As it is basically a binarization work, two clusters (*C=2*) are taken in account. The ink-strokes (object region) are converted into black and remaining background is converted into white. Euclidean distance function is used in this *FCM* clustering. In this way, we identify the ink-stroked (foreground) region consisting of texts and non-text doodles in Tagore's manuscript (*fig.2*).

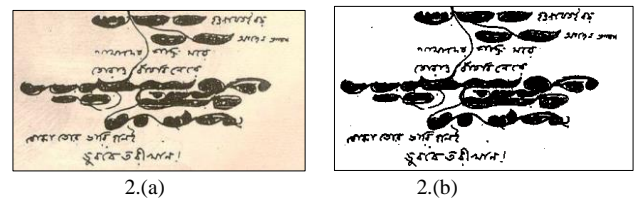

2.(a)      2.(b)

Fig. 2. Tagore's Manuscript: (a) Original, (b) Object region after using FCM

### B. Separation of Texts and Non-Texts

The non-texts in a Tagore's manuscript consist of the doodles. These doodles are mostly drawn by the same ink of the textual portion. The textual region may overlap or touch the doodles. Sometimes, textual portions are struck-through by a pen. So, the separation of texts and doodles is challenging task. Here we use basic features of texts and doodles to do the separation: texts are made with elongated, thin and curvy lines, and a doodle is relatively large, well connected and contains dense component.

The connected component analysis is used to find the major portion of a doodle, i.e. a large sized component is more likely to be a part of a doodle. Each *5X5* window is examined and the number of black pixels ($C_{black}$) are counted; if $C_{black} > T1$, where *T1* is a threshold, then it is considered as a dense component and marked as a doodle portion. The horizontal / vertical / diagonal struck-out lines are removed by checking degree of continuity. The details of this separation task can be found in our previous work [6].

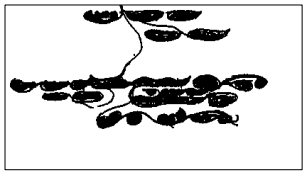

Fig. 3. Extracted doodle from fig.2.

### C. Text Line Identification

The objective of this paper is identification of text lines. On the way of handwritten character recognition, it is important to find the textual region and to detect the text-lines. We use the basic feature of text line, i.e. generally it is elongated in a particular direction (suppose, horizontal) than other (vertical). There are less numbers (<*threshold T2*) of text-pixels in a certain width stretched at few numbers of rows continuously, these are the inter text-line white-gaps.

After text/ non-text separation (§.II.*B*), we deal with the textual portion (say, image $I_{text}$: *fig.4.a*). The Gaussian filter is used on $I_{text}$ for smoothing the text region and it is scanned from left to right by examining *1XW* block (*W*: block width). In this block, if the number of black text-pixels (*N*) is greater than a threshold (*T>0*), i.e. *N>T>0*, all the pixels of that block are converted into black (*fig.4.b*). It is one kind of *black run length smearing*, and it is done to get an enhanced histogram (*fig.4.c*). To produce a *horizontal histogram* (HH), the number of black (object) pixels of each row is counted and plotted in *x*-axis with row index in *y*-axis. The *peak* of a HH denotes higher number of black pixels in a row i.e. text line, and similarly a *valley* denotes the inter text-line white gap (with a small number of black pixels). Then connected component (CC) analysis is used. As our text/ non-text separator does not produce *100%* correct result, some portions of non-texts may be mixed with text portion. These mixed non-texts are removed by using morphological *'clean'* [5] operation (eradicating isolated point) and eliminating small smear component (*fig.4.d*). Due to smearing, each component denotes a separate line (*fig.4.f*) in the favorable cases. But sometimes when two lines are connected, HH helps to analyze. In HH it is known that which rows have a larger amount of black pixels (near to peak), i.e. those rows are mostly in the text-lines, otherwise (near to valley) those are in the inter-line gaps. The drastic change from valley to peak (or, from peak to valley) is noticed in a HH (of each connected component), and an overlapping region in text-lines is found. An overlapped text-line is found by calculating the height of those components, as the component containing overlapped text-lines has more height than others. Then, those two connected text-lines are separated by analyzing the minima and drastic change in HH (*fig.6*).

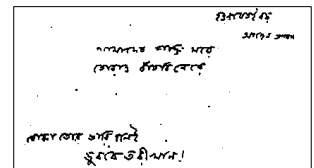
4.(a)

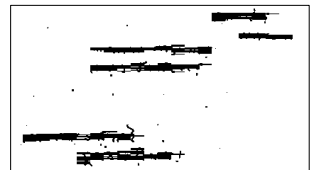 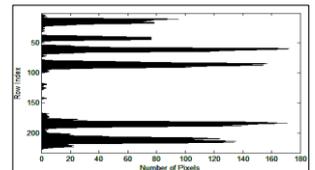
4.(b) 4.(c)

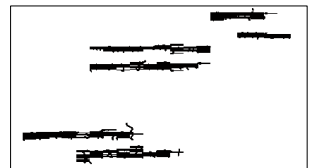 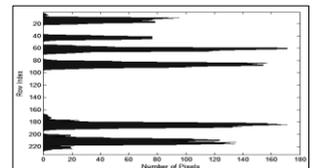
4.(d) 4.(e)

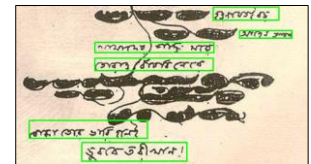
4.(f)

Fig.4. **(a)** Text region of fig.2., **(b)** Black run length smearing of 4.(a), **(c)** HH of 4.(b), **(d)** Larger CC of 4.(b), **(e)** HH of 4.(d), **(f)** Text line identified and marked by green boxes.

### III. EXPERIMENTAL RESULTS AND DISCUSSION

We implemented our proposed method in MATLAB [5]. In order to assess the correctness of our proposed method, we take 45 numbers of Tagore's manuscript written in English and Bengali.

Some experimental results are shown in *fig.5(a)-(d)* and *fig.6*.

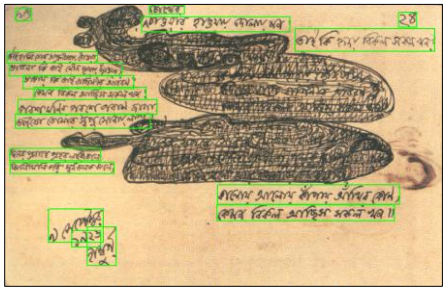

5.(a)

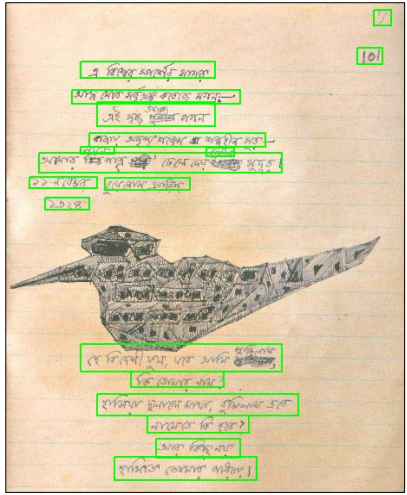

5.(b)

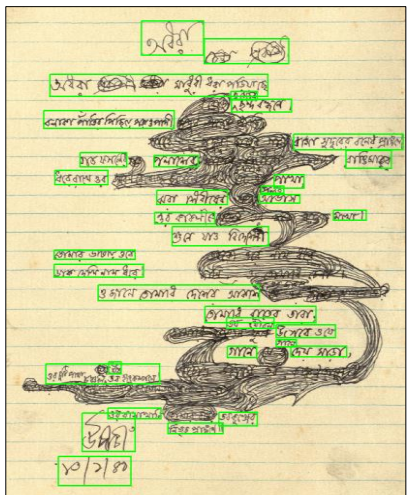

5.(c)

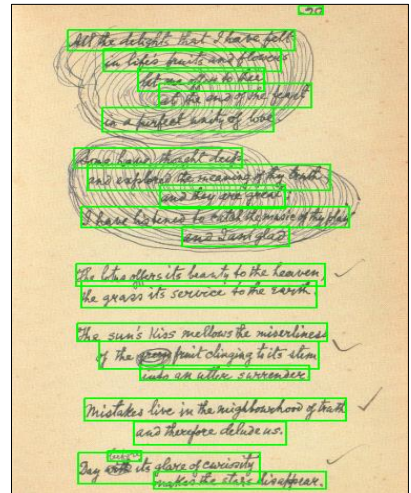

5.(d)

Fig.5. **(a)-(d)** Text line identification and marked by green boxes.

*Fig.5.(d)* shows that our proposed method works on English script and identifies struck-out text region.

In fig.6, we deal with the overlapping lines using HH. The valley of HH indicates minimum number of object pixels in the row. For a component (*fig.6.b*), we find the drastic change in HH from valley to peak and marked as an overlapping region. The height of a component gives more surety of overlapping as overlapped textual component's height is more than non-overlapped one.

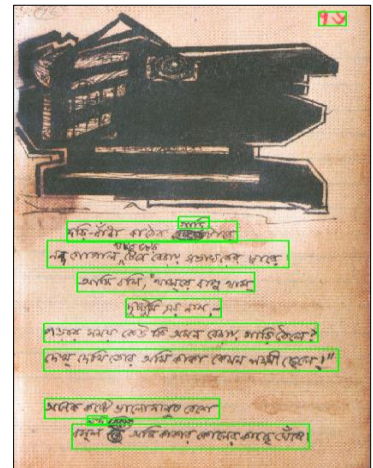

6.(a)

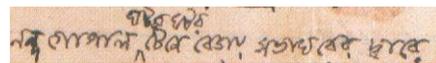

6.(b)

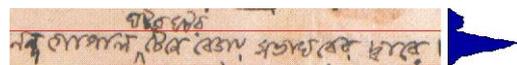

6.(c)

Fig.6. Manuscript of Tagore's poem ('Natun Srota'), *Parisesh*, 19 Aug., 1927: **(a)** Text line marked by green boxes, **(b)** Two overlapping lines of 6.(a), **(c)** Separation of overlapping line is shown by red line, beside that HH is shown.

For performance evaluation we find *Precision (P), Recall (R)* and *F-Measure (FM)* as their usual meaning. We measure *P*, *R* and *FM* for each manuscript, and finally average (arithmetic mean) them (shown in *table.1*).

For a manuscript *true positive* (TP), *false positive* (FP) and *false negative* (FN) are defined as follows:

TP: # text-lines those are correctly classified,
FP: # text-lines those are incorrectly classified (unexpected),
FN: # text-lines those are not classified (missing result).

*P=TP/(TP+FP)*, *R=TP/(TP+FN)*, *FM=(2.P.R)/(P+R)*.

**Table 1.** Evaluation of Results

| P (%) | R (%) | FM (%) |
|---|---|---|
| 87.39 | 91.57 | 89.41 |

Our method works on skewed document which is tilted slightly (skew angle ≈ $45^o$) from horizon (*fig.5.b*, bottom left). But, it fails in a case where the writings are in vertical direction (skew angle ≈ $90^o$). In this scenario, we rotate the document manually and fed into the system. If a manuscript contains texts in different directions (*fig.7*), manually we cut the manuscript in different directions before identification of text-lines.

In Bengali script, there are some characters made of disjoint components (such as in English: 'i' and 'j') shown in *fig.8* (printed Bengali script), each character is separated by a box. In Tagore's manuscript most of the times, bottom dots of the characters shown in first four boxes in *fig.8*, become very closed to the lower text-line instead of the upper line (*fig.9*). Our system fails to identify these dots of upper line, and these become part of the bottom line. However, it gives more challenge to work on Tagore's manuscript.

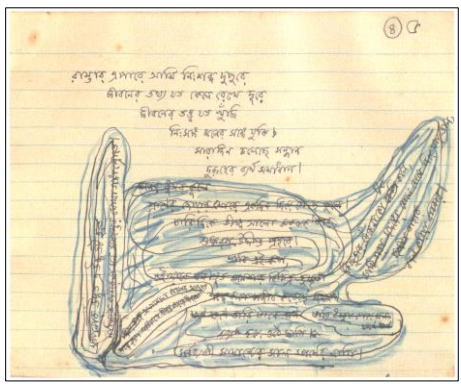
Fig.7. Tagore's manuscript (*Nabajatok*, 4 May, 1939): Text lines in different directions with struck-out lines.

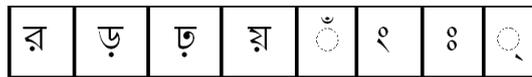
Fig.8. Bengali printed characters with disjoint components.

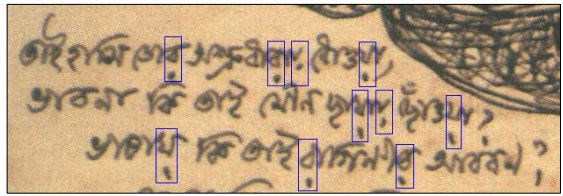
Fig.9. Tagore's manuscript contains special characters of *fig.8*, manual marking by blue boxes.

To the best of our knowledge, it is the first work on text line identification in Tagore's manuscript. We did not get any other reference/ work to do a comparative study.

IV. CONCLUSION

A text line identifier for Tagore's manuscript has been presented in this paper. We obtained 89.41% of F-measure on the basis of our experiment on the test cases. As Tagore's manuscripts constitute one of the most difficult cases, our method is expected to work with manuscripts of other notable writers.

Still there are many unpublished works of Tagore. To check the originality of some such manuscript, some handwriting recognition expert is called for. Our target is to make a computerized approach to know whether those are written by Tagore.

At present the system is semi-automatic. Our next endeavor will be to make the system automatic and more accurate.